%% file: main.tex
\documentclass{article}
\PassOptionsToPackage{numbers, compress}{natbib}
    \usepackage[preprint]{neurips_2022}

\usepackage[utf8]{inputenc} 
\usepackage[T1]{fontenc}    
\usepackage{hyperref}       
\usepackage{url}            
\usepackage{booktabs}       
\usepackage{amsfonts}       
\usepackage{nicefrac}       
\usepackage{microtype}      
\usepackage{xcolor}         
\usepackage{enumitem}
\usepackage{float}
\usepackage{etoolbox}
\makeatletter
\patchcmd{\@algocf@start}
  {-1.5em}
  {0pt}
  {}{}
\makeatother
\usepackage{multicol}
\usepackage{svg}

\usepackage{colortbl}

\usepackage{hyperref}
\usepackage{url}
\usepackage{bm}
\usepackage{amsmath}
\usepackage{tikz}
\usepackage{arydshln}
\usepackage[symbol]{footmisc}
\usepackage{footnote}
\usepackage{multirow}
\usepackage[T1]{fontenc}    
\usepackage{graphicx}
\usepackage{wrapfig}
\usepackage{wrapfig,lipsum,booktabs}

\usepackage{amssymb}
\usepackage{array}
\usepackage{arydshln}
\usepackage{setspace}
\usepackage{comment}

\hypersetup{colorlinks,citecolor={teal}}

\usepackage{caption}
\usepackage{subcaption}
\include{math_definition}

\usepackage{xcolor}
\definecolor{navyblue}{rgb}{0.0, 0.0, 0.5}
\definecolor{darkblue}{rgb}{0.0, 0.0, 0.55}
\hypersetup{
    colorlinks = true,
    citecolor=darkblue,
    linkcolor=red,
    urlcolor  = blue,
    anchorcolor = blue}

\usepackage{supertabular}
\usepackage{array,booktabs}
\usepackage{amsmath}
\usepackage{bm}
\usepackage{amsfonts}
\usepackage{multirow}

\usepackage{longtable}

\usepackage{amsmath}
\usepackage{amssymb}
\usepackage{mathtools}
\usepackage{amsthm}
\usepackage[normalem]{ulem}

\usepackage{amsfonts}
\usepackage{bm}

\DeclareMathOperator*{\argmin}{arg\,min}
\DeclareMathOperator*{\minimize}{minimize}

\DeclareMathOperator*{\diag}{diag}

\usepackage{eqparbox}
\usepackage{makecell}
\usepackage{threeparttable}

\usepackage{algorithm}
\usepackage[noend]{algorithmic}
\renewcommand{\algorithmiccomment}[1]{\hfill\eqparbox{COMMENT}{\footnotesize $\rhd$ #1}}

\usepackage[capitalize,noabbrev,nameinlink]{cleveref}

\theoremstyle{plain}

\theoremstyle{definition}

\theoremstyle{remark}

\usepackage[textsize=tiny]{todonotes}

\newcommand{\eat}[1]{}

\definecolor{Red}{rgb}{0.6,0,0}
\definecolor{Blue}{rgb}{0,0,0.8}
\definecolor{Green}{rgb}{0,0.4,0.7}
\definecolor{airforceblue}{rgb}{0.36, 0.54, 0.66}
\definecolor{ao(english)}{rgb}{0.0, 0.5, 0.0}
\definecolor{azure(colorwheel)}{rgb}{0.0, 0.5, 1.0}
\definecolor{crimson}{rgb}{0.86, 0.08, 0.24}
\definecolor{darkcerulean}{rgb}{0.03, 0.27, 0.49}
\definecolor{cobalt}{rgb}{0.0, 0.28, 0.67}
\definecolor{rosegold}{rgb}{0.72, 0.43, 0.47}
\definecolor{orange-red}{rgb}{1.0, 0.27, 0.0}
\definecolor{mountainmeadow}{rgb}{0.19, 0.73, 0.56}
\definecolor{malachite}{rgb}{0.04, 0.85, 0.32}
\definecolor{darkblue}{rgb}{0.0, 0.0, 0.55}

\definecolor{customblue}{rgb}{0.2, 0.35, 0.8}
\definecolor{customcolor}{gray}{0.}
\definecolor{gg}{gray}{0.9}
\definecolor{tg}{gray}{0.6}
\usepackage{hyperref}
\hypersetup{colorlinks=true}
\hypersetup{linktoc=all}
\hypersetup{citecolor=customblue}
\hypersetup{linkcolor=crimson}
\hypersetup{urlcolor=MidnightBlue}
\usepackage[all]{hypcap}

\creflabelformat{equation}{#2\textup{#1}#3}  

\newcommand{\bsy}{\boldsymbol}
\newcommand{\highlight}[1]{{\color{crimson}{#1}}}
\newcommand{\TBD}[1]{{\color{azure(colorwheel)}{#1}}} 
\newcommand{\modify}[1]{{\color{orange-red}{#1}}} 

\newcommand{\txtgray}[1]{{\color{tg}{#1}}}
\newcommand{\toxic}[1]{{\color{customcolor}{\textbf{#1}}}}

\newcounter{daggerfootnote}

\title{BiTAT: Neural Network Binarization \\with Task-dependent Aggregated Transformation}
\author{%
  Geon Park$^{1}$\thanks{Equal contribution.}\hspace{0.15in} 
  Jaehong Yoon$^{1}$\footnotemark[1] \hspace{0.15in} 
  Haiyang Zhang$^{3}$ \hspace{0.15in} 
  Xing Zhang$^{3}$\\ 
  \textbf{
  Sung Ju Hwang$^{1~2}$ \hspace{0.15in} 
  Yonina  C. Eldar$^{3}$}\\
  KAIST$^{1}$ \hspace{0.1in}
  AITRICS$^{2}$ \hspace{0.1in}
  Weizmann Institute of Science$^{3}$
}

\begin{document} 

\maketitle
\setlength{\textfloatsep}{5pt}

\input{sections/0_abstract}
\input{sections/1_introduction}
\input{sections/2_related_work}
\input{sections/3_motivation}
\input{sections/4_approach}

\input{sections/5_experiments}

\input{sections/6_conclusion}

\bibliography{reference}

\clearpage

\begin{enumerate}

\item For all authors...
\begin{enumerate}
  \item Do the main claims made in the abstract and introduction accurately reflect the paper's contributions and scope?
    \answerYes{}
  \item Did you describe the limitations of your work?
    \answerYes{See supplementary file.}
  \item Did you discuss any potential negative societal impacts of your work?
    \answerYes{See supplementary file.}
  \item Have you read the ethics review guidelines and ensured that your paper conforms to them?
    \answerYes{}
\end{enumerate}

\item If you are including theoretical results...
\begin{enumerate}
  \item Did you state the full set of assumptions of all theoretical results?
    \answerNA{}
        \item Did you include complete proofs of all theoretical results?
    \answerNA{}
\end{enumerate}

\item If you ran experiments...
\begin{enumerate}
  \item Did you include the code, data, and instructions needed to reproduce the main experimental results (either in the supplemental material or as a URL)?
    \answerYes{See supplementary materials.}
  \item Did you specify all the training details (e.g., data splits, hyperparameters, how they were chosen)?
    \answerYes{See supplementary file.}
\item Did you report error bars (e.g., with respect to the random seed after running experiments multiple times)?
\answerYes{}
\item Did you include the total amount of compute and the type of resources used (e.g., type of GPUs, internal cluster, or cloud provider)?
    \answerYes{See supplementary file.}
\end{enumerate}

\item If you are using existing assets (e.g., code, data, models) or curating/releasing new assets...
\begin{enumerate}
  \item If your work uses existing assets, did you cite the creators?
    \answerYes{}
  \item Did you mention the license of the assets?
    \answerNo{}
  \item Did you include any new assets either in the supplemental material or as a URL?
    \answerNA{}
  \item Did you discuss whether and how consent was obtained from people whose data you're using/curating?
    \answerNo{}
  \item Did you discuss whether the data you are using/curating contains personally identifiable information or offensive content?
    \answerNo{}
\end{enumerate}

\item If you used crowdsourcing or conducted research with human subjects...
\begin{enumerate}
  \item Did you include the full text of instructions given to participants and screenshots, if applicable?
    \answerNA{}
  \item Did you describe any potential participant risks, with links to Institutional Review Board (IRB) approvals, if applicable?
    \answerNA{}
  \item Did you include the estimated hourly wage paid to participants and the total amount spent on participant compensation?
    \answerNA{}
\end{enumerate}

\end{enumerate}

\input{sections/7_appendix}

\end{document}

%% file: sections/0_abstract.tex
\begin{abstract}
Neural network quantization aims to transform high-precision weights and activations of a given neural network into low-precision weights/activations for reduced memory usage and computation, while preserving the performance of the original model. However, extreme quantization (1-bit weight/1-bit activations) of compactly-designed backbone architectures (e.g., MobileNets) often used for edge-device deployments results in severe performance degeneration. This paper proposes a novel Quantization-Aware Training (QAT) method that can effectively alleviate performance degeneration even with extreme quantization by focusing on the inter-weight dependencies, between the weights within each layer and across consecutive layers. To minimize the quantization impact of each weight on others, we perform an orthonormal transformation of the weights at each layer by training an input-dependent correlation matrix and importance vector, such that each weight is disentangled from the others. Then, we quantize the weights based on their importance to minimize the loss of the information from the original weights/activations. We further perform progressive layer-wise quantization from the bottom layer to the top, so that quantization at each layer reflects the quantized distributions of weights and activations at previous layers.  We validate the effectiveness of our method on various benchmark datasets against strong neural quantization baselines, demonstrating that it alleviates the performance degeneration on ImageNet and successfully preserves the full-precision model performance on CIFAR-100 with compact backbone networks.
\end{abstract}


%% file: sections/1_introduction.tex
\vspace{-.2in}
\section{Introduction}

Over the past decade, deep neural networks have achieved tremendous success in solving various real-world problems, such as 
image/text generation~\cite{creswell2018generative,karras2021alias}, unsupervised representation learning~\cite{gidaris2018unsupervised,chen2020simple,zbontar2021barlow}, and multi-modal training~\cite{su2019vl,zareian2021open,radford2021learning}. Recently, network architectures that aim to solve target tasks are becoming increasingly larger, based on the empirical observations of their improved performance. However, as the models become larger, it is increasingly difficult to deploy them on resource-limited edge devices with limited memory and computational power. 
Therefore, many recent works focus on building resource-efficient deep neural networks to bridge the gap between the scale of deep neural networks and actual permissible computational complexity/memory-bounds for on-device model deployments. Some of these works consider designing computation- and memory-efficient modules for neural architectures, while others focus on compressing a given neural network by either pruning its weights~\cite{dai2018compressing,he2020learning,lin2020hrank,yoon2017combined} or reducing the bits used to represent the weights and activations~\cite{bulat_high-capacity_2021,dbq,li_brecq_2021}. The last approach, \emph{neural network quantization}, is beneficial for building on-device AI systems since the edge devices oftentimes only support low bitwidth-precision parameters and/or operations. However, it inevitably suffers from the non-negligible forgetting of the encoded information from the full-precision models. Such loss of information becomes worse with extreme quantization into binary neural networks with 1-bit weights and 1-bit activations~\cite{bulat_high-capacity_2021,zhuang2019structured,rastegari2016xnor,qin2020binary}. 

How can we then effectively preserve the original model performance even with extremely low-precision deep neural networks? To address this question, we focus on the somewhat overlooked properties of neural networks for quantization: the weights in a layer are highly correlated with each other and weights in the consecutive layers. Quantizing the weights will inevitably affect the weights within the same layer, since they together comprise a transformation represented by the layer. Thus, quantizing the weights and activations at a specific layer will adjust the correlation and relative importance between them. Moreover, it will also largely impact the next layer that directly uses the output of the layer, which together comprise a function represented by the neural network.

\input{materials/tables/2_categorization}

Despite their impact on neural network quantization, such inter-weight dependencies have been relatively overlooked. As shown in \Cref{tab:categorization}~\highlight{Right}, although BRECQ~\cite{li_brecq_2021} addresses the problem by considering the dependency between filters in each block, it is limited to the Post-Training Quantization (PTQ) problem, which suffers from inevitable information loss, resulting in inferior performance. The most recent Quantization-Aware Training (QAT) methods~\cite{dbq,liu_reactnet_2020} are concerned with obtaining quantized weights by minimizing quantization losses with parameterized activation functions, disregarding cross-layer weight dependencies during training process. To the best of our knowledge, no prior work explicitly considers dependencies among the weights for QAT.

To tackle this challenging problem, we propose a new QAT method, referred to as Neural Network \textbf{Bi}narization with \textbf{T}ask-dependent \textbf{A}ggregated \textbf{T}ransformation (\textbf{BiTAT}), as illustrated in \Cref{tab:categorization}~\highlight{Left}. Our method sequentially quantizes the weights at each layer of a pre-trained neural network based on chunk-wise input-dependent weight importance by training orthonormal dependency matrices and scaling vectors. While quantizing each layer, we fine-tune the subsequent full-precision layers, which utilize the quantized layer as an input for a few epochs while keeping the quantized weights frozen. we aggregate redundant input dimensions for transformation matrices and scaling vectors, significantly reducing the computational cost of the quantization process. Such consideration of inter-weight dependencies allows our BiTAT algorithm to better preserve the information from a given high-precision network, allowing it to achieve comparable performance to the original full-precision network even with extreme quantization, such as binarization of both weights and activations. The main contributions of the paper can be summarized as follows:
\begin{itemize}
\item We demonstrate that weight dependencies within each layer and across layers play an essential role in preserving the model performance during quantized training. 
\item We propose an input-dependent quantization-aware training method that binarizes neural networks. We disentangle the correlation in the weights from across multiple layers by training rotation matrices and importance vectors, which guides the quantization process to consider the disentangled weights' importance.
\item We empirically validate our method on several benchmark datasets against state-of-the-art neural network quantization methods, showing that it significantly outperforms baselines with the compact neural network architecture.
\end{itemize}

%% file: materials/tables/2_categorization.tex
\begin{figure*}[t]
\small
\hspace{-0.05in}
\begin{minipage}{0.65\linewidth}
\centering
\includegraphics[width=\linewidth]{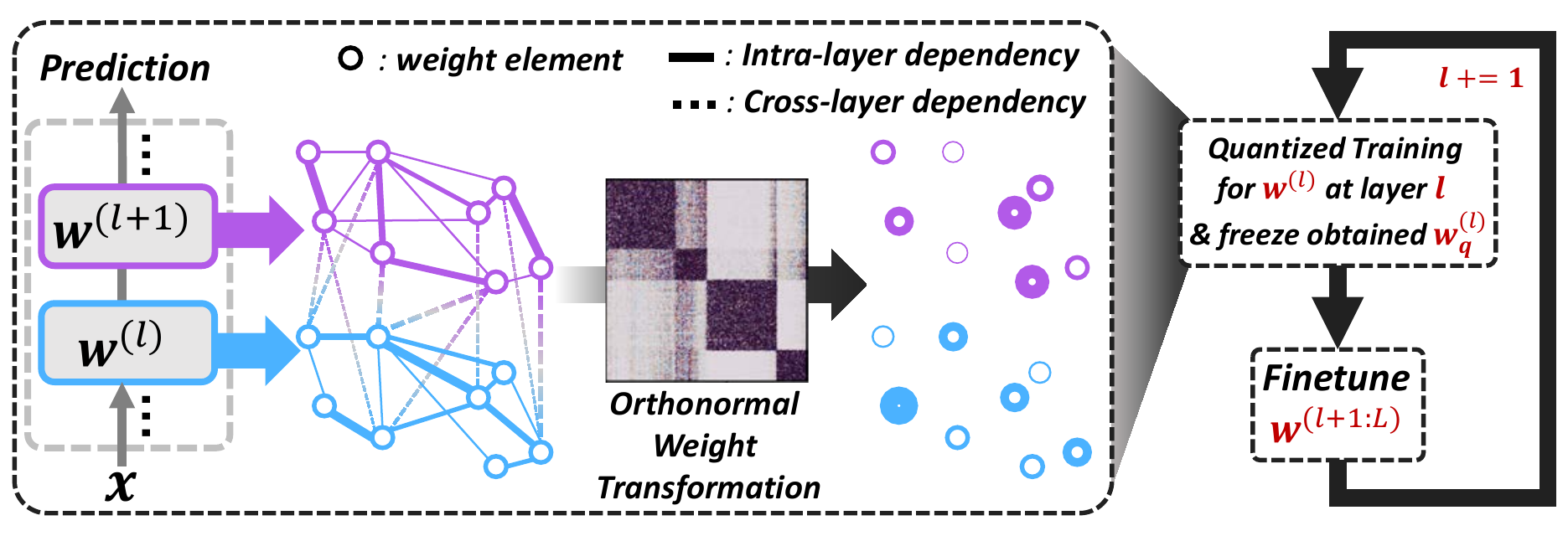}
\end{minipage}\hspace{0.05in}
\begin{minipage}{0.35\linewidth}
\centering
\resizebox{\textwidth}{!}{
\setlength\tabcolsep{2pt}
\centering
\begin{tabular}{lccccc}
\toprule
\textsc{Method} & \thead{BRECQ \\ \cite{li_brecq_2021}} & \thead{DBQ \\ \cite{dbq}} &\thead{ReActNet \\  \cite{liu_reactnet_2020}} & \textbf{Ours}\\
\midrule
\textsc{$\text{bit}_{w}$ / $\text{bit}_{a}$}  & $2/4$           & $4/8$ & $1/1$          & $1/1$          \\
\textsc{correlation}     & block &N/A &N/A & block \\
\textsc{task-based $Q$}     & $\checkmark$  & $\times$ & $\times$ & $\checkmark$ \\
\textsc{structured} & node  & $\times$        & $\times$        & dynamic     \\ 
\textsc{approach}         & PTQ$^1$       & QAT$^2$ & QAT& QAT          \\ \midrule
\textsc{T1@ImgNet}    & 66.60\%     & 70.5\% & 68.26\%         & \textbf{{68.51\%}}          \\
\textsc{FLOPs} $\times 10^7$    & 3.31    & 3.60 & \textbf{1.2}         & \textbf{1.2}          \\
\bottomrule
\end{tabular}
}
\begin{tablenotes}
\item[1] \hspace{-0.1in}\scriptsize{$^1$Post-training Quantization}
\item[2] \hspace{-0.1in}\scriptsize{$^2$Quantization-aware Training}
\end{tablenotes}
\end{minipage}
\vspace{-0.1in}
\caption{\small \textbf{Left: An Illustration of our proposed method.} 
Weight elements in a layer is highly correlated to each other along with the weights in other layers. Our BiTAT sequentially obtains quantized weights of each layer based on the importance of disentangled weights to others using a trainable orthonormal rotation matrix and importance vector.
\textbf{Right:} Categorization of relevant and strong quantization methods to ours.  \label{tab:categorization}}
\end{figure*}


%% file: sections/2_related_work.tex
\vspace{-0.05in}
\section{Related Work}\label{sec:related_work}
\vspace{-0.05in}

\paragraph{Minimizing the quantization error.} 
Quantization methods for deep neural networks can be broadly categorized into several strategies~\citep{bnn_servey}. We first introduce the methods by minimizing the quantization error. Many existing neural quantization methods aim to minimize the weight/activation discrepancy between quantized models and their high-precision counterparts. XNOR-Net~\citep{xnornet} aims to minimize the least-squares error between quantized and full-precision weights for each output channel in layers. DBQ~\citep{dbq} and QIL~\citep{qil} perform layerwise quantization with parametric scale or transformation functions optimized to the task. Yet, they quantize full-precision weight elements regardless of the correlation between other weights. While TSQ~\citep{twostep} and Real-to-Bin~\cite{realtobin} propose to minimize the $\ell_2$ distance between the quantized activations and the real-valued network's activations by leveraging intra-layer weight dependency, they do not consider cross-layer dependencies.
Recently, BRECQ~\citep{li_brecq_2021} and the work in a similar vein on post-training quantization~\cite{nagel_up_2020} consider the interdependencies between the weights and the activations by using a Taylor series-based approach. However, calculating the Hessian matrix for a large neural network is often intractable, and thus they resort to strong assumptions such as small block-diagonality of the Hessian matrix to make them feasible. BiTAT solves this problem by training the dependency matrices alongside the quantized weights while grouping similar weights together to reduce the computational cost.

\paragraph{Modifying the task loss function.}
A line of methods aims to achieve better generalization performance during quantization by taking sophisticatedly-designed loss functions. BNN-DL~\cite{regactdist} adds a distributional loss term that enforces the distribution of the weights to be quantization-friendly. Apprentice~\cite{mishra2018apprentice} uses knowledge distillation (KD) to preserve the knowledge of the full-precision teacher network in the quantized network. However, such methods only put a constraint on the distributional properties of the weights, not the dependencies and the values of the weight elements. CI-BCNN~\cite{Wang_2019_CVPR} parameterizes bitcount operations by exploring the interaction between output channels using reinforcement learning and quantizes the floating-point accumulation in convolution operations based on them. However, reinforcement learning is expensive, and it still does not consider cross-layer dependencies.

\paragraph{Reducing the gradient error.}
Bi-Real Net~\citep{liu_bi-real_2018} devises a better gradient estimator for the sign function used to binarize the activations and a magnitude-aware gradient correction method. It further modifies the MobileNetV1's architecture to better fit the quantized operations.
ReActNet~\citep{liu_reactnet_2020} achieves state-of-the-art performance for binary neural networks by training a generalized activation function for compact network architecture used in~\cite{liu_bi-real_2018}. However, the quantizer functions in these methods conduct element-wise unstructured compression without considering the change in other correlated weights during quantization training. This makes the search process converge to the suboptimal solutions since task loss is the only guide for finding the optimal quantized weights, which is often insufficient for high-dimensional and complex architectures. However, our proposed method can obtain a better-informed guide that compels the training procedure to spend more time searching in areas that are more likely to contain high-performing quantized weights. 

%% file: sections/3_motivation.tex
\section{Weight Importance for Quantization-aware Training}\label{sec:motivation}
We first introduce the problem of Quantization-Aware Training (QAT) in \Cref{subsec:statement} and show that the dependency among the neural network weights plays a crucial role in preserving the performance of a quantized model obtained with QAT in \Cref{subsec:input-dependent}. We further show that the dependency between consecutive layers critically affects the performance of the quantized model in \Cref{subsec:cross-layer}.

\subsection{Problem Statement}\label{subsec:statement}
We aim to quantize a full-precision neural network into a binary neural network (BNN), where the obtained quantized network is composed of binarized $1$-bit weights and activations, which preserves the performance of the original full-precision model. 
Let $f(\cdot;\mathcal{W})$ be a $L$-layered neural network parameterized by a set of pre-trained weights $\mathcal{W}=\{\bm{w}^{(1)}, \dots, \bm{w}^{(L)}\}$, where $\bm{w}^{(l)}\in\mathbb{R}^{d_{l-1}\times d_l}$ is the weight at layer $l$ and $d_0$ is the dimensionality of the input. Given a training dataset $\mathcal{X}$ and corresponding labels $\mathcal{Y}$, existing QAT methods~\cite{xnornet, dbq, qil, bethge_meliusnet_2020, yamamoto_learnable_2021, park_profit_2020} search for optimal quantized weights by solving for the optimization problem that can be generally described as follows:
\begin{align}
\underset{\mathcal{W},\bsy\phi}{\minimize}~~\mathcal{L}_{task}\left(f\left(\mathcal{X}; Q\left(\mathcal{W}; \bsy\phi\right)\right),\mathcal{Y}\right),
\end{align}
where $\mathcal{L}_{task}$ is a standard task loss function, such as cross-entropy loss, and $Q(\cdot; \bsy\phi)$ is the weight quantization function parameterized by $\bsy\phi$ which transforms a real-valued vector to a discrete, binary vector. 
The quantization function used in existing works typically minimize loss terms based on the Mean Squared Error (MSE) between the full-precision weights and the quantized weights at each layer:
\begin{align}\label{eq:q}
Q(\bm w) := \alpha^* \bm b^*, \text{~~~~ where~} \alpha^*, \bm b^* = \argmin_{\alpha \in \mathbb R, \bm b \in \{-1, 1\}^{m}} \left\Vert \bm w - \alpha \bm b \right\Vert_2^2,
\end{align}
\input{materials/figures/3_scatter} 
\noindent where $m$ is the dimensionality of the target weight. For inference, QAT methods use $\bm w_q = Q(\bm w)$ as the final quantized parameters.
They iteratively search for the quantized weights based on the task loss with a stochastic gradient descent optimizer, where the model parameters converge into the ball-like region around the full-precision weights $\bm w$.

However, the region around the optimal full-precision weights may contain suboptimal solutions with high errors. We demonstrate such inefficiency of the existing quantizer formulation through a simple experiment in  \Cref{fig:scatter}. Suppose we have three input points, $\bm x_1, \bm x_2$, and $\bm x_3$, and full-precision weights $\bm w$.
Quantized training of the weight using \Cref{eq:q} successfully reduces MSE between the quantized weight and the full-precision, but the task prediction loss using $\bm w_q$ is nonetheless very high. 

We hypothesize that the main source of error comes from the independent application of the quantization process to each weight element. However, neural network weights are not independent, but highly correlated and thus quantizing a set of weights will largely affect the others. Moreover, after quantization, the relative importance among weights could also largely change. Both factors lead to high quantization errors in the pre-activations. On the other hand, our proposed QAT method \emph{BiTAT}, described in \Cref{sec:method}, achieves a quantized model with much smaller MSE. This results from the consideration of the inter-weights dependencies, which we describe in the next subsection.

\subsection{Disentangling Weight Dependencies via Input-dependent Orthornormal Transformation}
\label{subsec:input-dependent}

\input{materials/figures/4_motivation_validation}
How can we then find the low-precision subspace, which contains the best-performing quantized weights on the task, by exploiting the inter-weight dependencies? The properties in the input distribution give us some insights into this question.
Let us consider a task composed of centered $N$ training samples \{$\bm x_1, \dots, \bm x_N\}= \mathcal X\in \mathbb R^{N\times d_0}$. We can obtain principal components of the training samples $\bm v_1, \dots, \bm v_{d_0} \in \mathbb R^{d_0}$ and the corresponding coefficients $\lambda_1, \dots, \lambda_{d_0} \geq 0$, in a descending order. 
Let us further suppose that we optimize a single-layered neural network parameterized by $\bm w^{(1)}$. Neurons corresponding to the columns of $\bm w^{(1)}$ are oriented in a similar direction to the principal components with higher variances (i.e., $\bm v_i$ than $\bm v_j$, where $i<j$) that is much more likely to get activated than the others. We apply a change of basis to the column space of the weight matrix $\bm w^{(1)}$ with the bases $(\bm v_1, \dots, \bm v_{d_0})$:
\begin{align}
    \bm V \widetilde{\bm{w}}^{(1)} &= \bm w^{(1)}\\
    \widetilde{\bm{w}}^{(1)} &= {\bm V}^\top \bm w^{(1)}, \label{eq:tilde}
\end{align}
where $\bm V = [\bm v_1 \ |\ \cdots\ |\ \bm v_{d_0}] \in \mathbb R^{d_0 \times d_0}$ is an orthonormal matrix. The top rows of the transformed weight matrix $\widetilde{\bm{w}}^{(1)}$ will contain more important weights, whereas the bottom rows will contain less important ones. Therefore, the accuracy of the model will be more affected by the perturbations of the weights at top rows than ones at the bottom rows. Note that this transformation can also be applied to the convolutional layer by ``unfolding'' the input image or feature map into a set of patches, which enables us to convert the convolutional weights into a matrix (The detailed descriptions of the orthonormal transformations for convolutional layers is provided in the \highlight{supplementary file}). 

We can also easily generalize the method to multi-layer neural networks, by taking the inputs for the $l$-th layer as the ``training set'', assuming that all of the previous layer's weights are fixed, as follows:
\begin{align}
    \left\{ \bm x^{(l)}_i = \sigma\left( {\bm w^{(l)}} ^\top \bm x^{(l-1)}_i \right) \right\}_{i=1}^N,
\end{align}
where $\sigma(\cdot)$ is the nonlinear transformation defined by both the non-linear activations and any layers other than linear transformation with the weights, such as a average-pooling or Batch Normalization. Then, we straightforwardly obtain the change-of-basis matrix $V^{(l)}$ and $\bm s^{(l)}$ for layer $l$. The impact of transformed weights is shown in \Cref{fig:motiv}. We compute the principal components of each layer in the initial pre-trained model and measure the test accuracy when adding the noise to the top-5 highest-variance \emph{(dashed red)} or lowest-variance components \emph{(dashed blue)} per layer.
While a model with perturbed high-variance components degenerates the performance as the noise scale increases, a model with perturbed low-variance components consistently obtains high performance even with large perturbations. This shows that preserving the important weight components that correspond to high-variance input components is critically important for effective neural network quantization that can preserve the loss of the original model.


\subsection{Cross-layer Weight Correlation Impacts Model Performance}\label{subsec:cross-layer}
So far, we only described the dependency among the weights within a single layer. However, dependencies between the weights across different layers also significantly impact the performance. To validate that, we perform layerwise sequential training from the bottom layer to the top. At the training of each layer, the model computes the principal components of the target layer and adds noise to its top-5 high/low components. As shown in \Cref{fig:motiv}, progressive training with the low-variance components~\emph{(solid blue)} achieves significantly improved accuracy over the end-to-end training counterpart~\emph{(dashed blue)} with a high noise scale, which demonstrates the beneficial effect of modeling weight dependencies in earlier layers. We describe further details in the \highlight{supplementary file}. 

\eat{\begin{align*}
    V^{(block)} = \left[
        \setlength{\tabcolsep}{0pt} 
        {\renewcommand{\arraystretch}{0}
        \begin{tabular}{ccccc}
            &&& \\
            &\framebox{\begin{minipage}[c][20pt][c]{20pt}\centering$V^{(1)}$\end{minipage}} && \dots& \\
            &&\begin{minipage}[c][15pt][c]{15pt}\centering$\ddots$\end{minipage}&&\\
            &\dots && \framebox{\begin{minipage}[c][30pt][c]{30pt}\centering$V^{(k)}$\end{minipage}}& \\
            &&&
        \end{tabular}
        }
    \right]
\end{align*}
}

%% file: materials/figures/3_scatter.tex
\begin{wrapfigure}{r}{0.45\textwidth}
    \vspace{-0.2in}
    \small
    \centering
    \includegraphics[width=\linewidth]{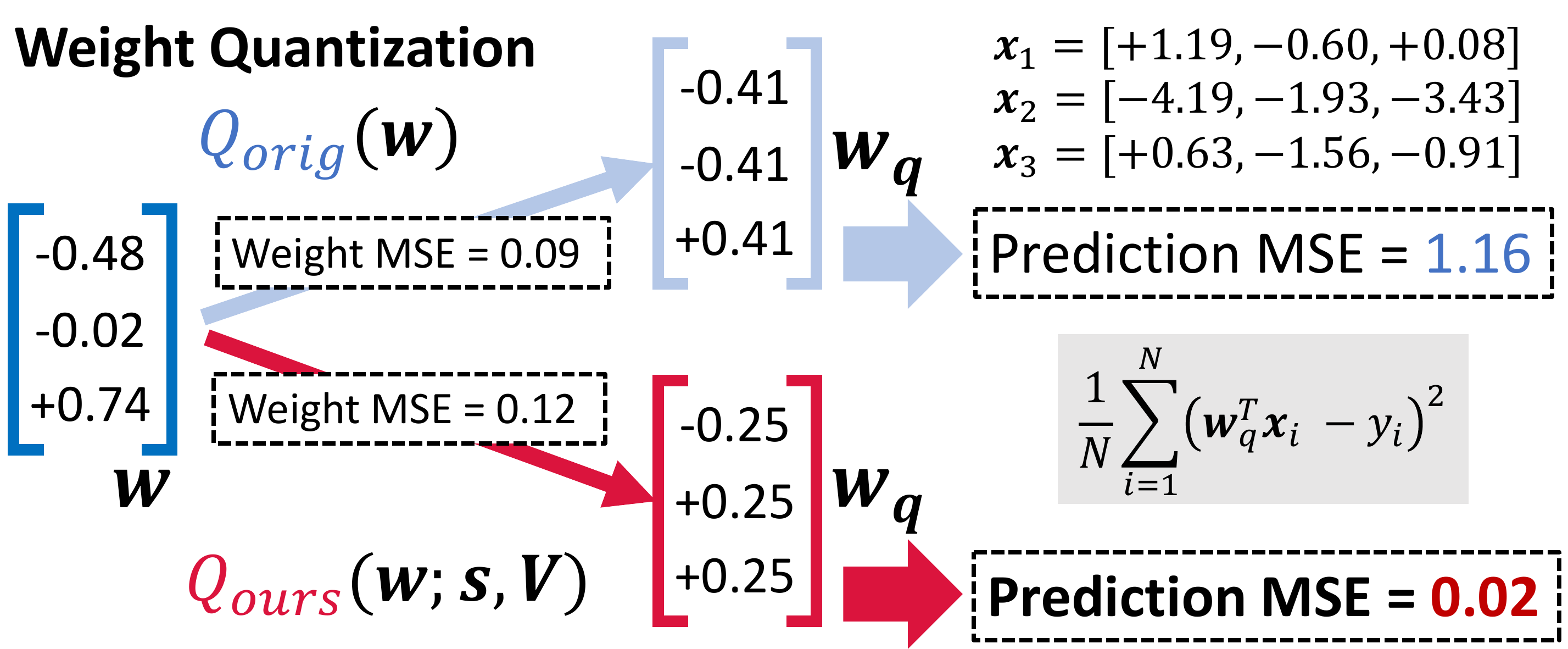}
    \vspace{-0.15in}
    \caption{\small A simple experiment that cross-layer weight correlation is critical to find well-performing quantized weights during QAT.}
    \label{fig:scatter}
    \vspace{-0.2in}
\end{wrapfigure} 

%% file: materials/figures/4_motivation_validation.tex
\begin{wrapfigure}{r}{0.3\textwidth}
    \centering
    \vspace{-0.2in}
    \includegraphics[width=\linewidth]{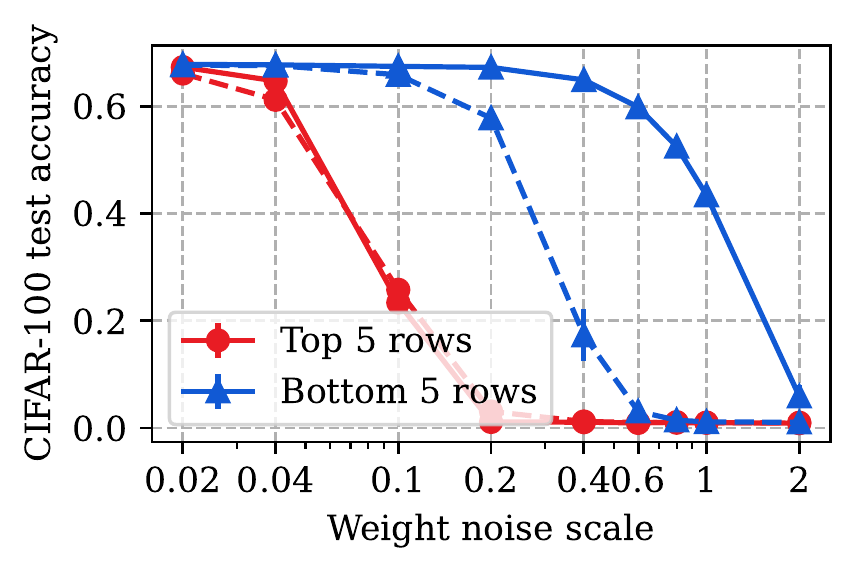}
    \vspace{-0.2in}
    \caption{\small \textbf{Solid lines:} Test accuracy of a MobileNetV2 model on CIFAR-100 dataset, after adding Gaussian noise to the top 5 rows and the bottom 5 rows of $\bm{\hat{w}}^{(l)}$ for all layers, considering the dependency on the lower layers. \textbf{Dashed lines:} Not considering the dependency on the lower layers. The x axis is in log scale.}
    \vspace{-0.4in}
    \label{fig:motiv}
\end{wrapfigure} 

%% file: sections/4_approach.tex
\section{Task-dependent Weight Transformation for Neural Network Binarization} \label{sec:method}

Our objective is to train binarized weights $\bm w_q$ given pre-trained full-precision weights. We effectively mitigate performance degeneration from the loss of information incurred by binarization by focusing on the inter-weight dependencies within each layer and across consecutive layers. We first reformulate the quantization function $Q$ in \Cref{eq:q} with the weight correlation matrix $\bm V$ and the importance vector $\bm s$ so that each weight is disentangled from the others while allowing larger quantization errors on the unimportant disentangled weights:
\begin{align}\label{eq:ourq}
    Q(\bm w;\bm s, \bm V) = \argmin_{\bm w_q \in \mathbb Q} \left\Vert \diag(\bm s)\left( {\bm V}^\top\otimes \bm w - {\bm V}^\top\otimes \bm w_q \right) \right\Vert^2_F + \gamma \left\Vert \bm w_q \right\Vert_1,
\end{align}
where $\bm s \in \mathbb R^{d_0}$ is a scaling term that assigns different importance scores to each row of $\hat{\bm{w}}$. We denote $\mathbb Q = \{ \bm \alpha \odot \bm b :  \bm\alpha \in \mathbb{R}^{d_{out}}, \bm{b} \in \{-1, 1\}^{d_{in}\times d_{out}} \}$ as the set of possible binarized values for $\bm w \in \mathbb{R}^{d_{in}\times d_{out}}$ with a scalar scaling factor for each output channel. 
The operation $\otimes$ denotes permuted matrix multiplication with index replacement, where we detail the computation in the following subsection.
We additionally include $\ell_1$ norm adjusted by a hyperparameter $\gamma$. At the same time, we want our quantized model to minimize the empirical task loss (e.g., cross-entropy loss) for a given dataset. Thus we formulate the full objective in the form of a bilevel optimization problem to find the best quantized weights which minimizes the task loss by considering the cross-layer weight dependencies and the relative importance among weights:
\begin{align} \label{eq:bilevel}
\begin{split}
    \bm w^*, \bm s^*, \bm V^* =& \argmin_{\bm w, \bm s, \bm V} \mathcal L_{task} \left( f\left(\mathcal X;\bm w_q\right), \mathcal Y \right),     \text{~~~~where } \bm w_q = Q(\bm w;\bm s, \bm V).
\end{split}
\end{align}
After the quantized training, the quantized weights $\bm w_q^*$ are determined by $\bm w_q^* = Q(\bm w^*; \bm s^*, \bm V^*)$. 

In practice, directly solving the above bilevel optimization problem is impractical due to its excessive computational cost. We therefore instead consider the following relaxed problem:
\begin{align}
\begin{split}\label{eq:relaxed}
    \bm w^*, \bm s^*, \bm V^* = \argmin_{\bm w, \bm s, \bm V} \mathcal L_{task} \left( f(\mathcal X; \mathrm{sgn}(\bm{w})), \mathcal Y \right) +& \lambda\left\Vert \diag(\bm s) {\bm V}^\top \otimes\left( \bm w - \mathrm{sgn}(\bm{w}) \right) \right\Vert^2_F\\ +& \gamma \left\Vert \mathrm{sgn}(\bm{w}) \right\Vert_1,
\end{split}
\end{align}
where $\lambda$ is a hyperparameter to balance between the quantization objective and task loss. Since it is impossible to compute the gradients for discrete values in quantized weights, we adopt the straight-through estimator~\cite{ste} that is broadly used across QAT methods: $\mathrm{sgn}(\bm{w})$ indicates the sign function applied elementwise to $\bm{w}$. We follow \cite{liu_reactnet_2020} for the derivative of $\mathrm{sgn}(\cdot)$.
Finally, we obtain the desired quantized weights by $\bm w_q^* = \mathrm{sgn}(\bm{w^*})$.
In order to obtain the off-diagonal parts of the cross-layer dependency matrix $V$, we minimize \Cref{eq:relaxed} with respect to $\bm s$ and $\bm V$ to dynamically determine the values 
(we omit $\mathcal X$ and $\mathcal Y$ from the arguments of $\mathcal{L}_{train}(\cdot)$ for readability):
\begin{align} \label{eq:relaxed2}
\begin{split}
\mathcal{L}_{train}(\bm w,\bm s, \bm V) = \mathcal{L}_{task}\left(f\left(\mathcal{X}; \mathrm{sgn}(\bm{w})\right), \mathcal{Y}\right) +&
        \lambda\left\Vert \diag(\bm s) {\bm V}^\top \otimes\left(\bm w - \mathrm{sgn}(\bm{w})\right)\right\Vert_F^2\\ 
        +& \gamma \left\Vert \mathrm{sgn}(\bm{w}) \right\Vert_1 + Reg(\bm s, \bm V),
\end{split}
\end{align}
where $Reg(\bm s, \bm V) := \Vert \bm V \bm V^\top - \bm I \Vert^2 + |\sigma - \sum_i \log(s_i)|^2$ is a regulariztion term which enforces $\bm V$ to be orthogonal and keeps the scale of $\bm s$ constant. Here, $\sigma$ is the constant initial value of $\sum_i \log(s_i)$, which is a non-negative importance score. 

\input{materials/figures/main_figure}

\subsection{Layer-progressive Quantization with Block-wise Weight Dependency}\label{subsec:method}
While we obtain the objective function in \Cref{eq:relaxed2}, it is inefficient to perform quantization-aware training while considering the complete correlations of all weights in the given neural network. Therefore, we only consider cross-layer dependencies between only few consecutive layers (we denote it as a \textit{block}), and initialize $\bm s$ and $\bm V$ using Principal Component Analysis (PCA) on the inputs to those layers within each block.

\input{materials/figures/4_block}
Formally, we define a weight correlation matrix in a neural network block $V^{(block)} \in \mathbb R^{(\sum_{i=1}^k d_i) \times (\sum_{i=1}^k d_i)}$, where $k$ is the number of layers in a block, similarly to the block-diagonal formulation in~\cite{li_brecq_2021} to express the dependencies between weights across layers in the off-diagonal parts. We initialize $\bm s^{(l)}$ and in-diagonal parts $V^{(l)}$ by applying PCA on the input covariance matrix:
\begin{align}\label{eq:initalize}
    \bm s^{(l)} \leftarrow {(\bm{\lambda}^{(l)})}^{\frac 12},\quad \bm V^{(l)} \leftarrow \bm U^{(l)}, \quad\quad\text{ where }~ \bm U^{(l)} \bm \lambda^{(l)} (\bm U^{(l)})^\top := \frac{1}{N}\sum_{i=1}^N \bm o_i^{(l-1)}  {\bm o_i^{(l-1)}}^\top,
\end{align}
where $\bm o^{(l)}$ is the output of $l$-th layer and $\bm o^{(0)}=\bm x$. This allows the weights at $l$-th layer to consider the dependencies on the weights from the earlier layers within the same neural block, and 
we refer to this method as \textit{sequential quantization}. Then, instead of having one set of $\bm s$ and $V$ for each layer, we can keep the previous layer's $\bm s$ and $V$ and expand them. Specifically, when quantizing layer $l$ which is a part of the block that starts with the layer $m$, we first apply PCA on the input covariance matrix to obtain $\bm{\lambda}^{(l)}$ and $U^{(l)}$. We then expand the existing $\bm s^{(m:l-1)}$ and $V^{(m:l-1)}$ to obtain $\bm s \in \mathbb{R}^{D + d_{l-1}}$ and $\bm s \in \mathbb{R}^{D + d_{l-1}}$ as follows\footnote{$[\cdot]_i$ indicates the $i$-th element of the object inside the brackets.}:
{\small
\begin{align}
    [\bm s^{(m:l)}]_i := \begin{cases} 
        [\bm s^{(m:l-1)}]_i, & i \leq D, \\
        [(\bm{\lambda}^{(l)})^{\frac 12}]_{i - D}, & D < i,
    \end{cases} &\quad\quad\quad
    [\bm V^{(m:l)}]_{i,j} := \begin{cases} 
        [\bm V^{(m:l-1)}]_{i,j}, & i, j \leq D, \\
        [\bm U^{(l)}]_{i - D, j - D}, & D < i, j, \\
        0, & \text{otherwise},
    \end{cases}
\end{align}}
where $D = \sum_{i=m}^{l-2}d_i$, as illustrated in~\Cref{fig:ours}. The weight dependencies between different layers (i.e., off-diagonal areas) are trainable and zero-initialized. To enable the matrix multiplication of the weights with the expanded $\bm s$ and $\bm V$, we define the expanded block weights as follows\footnote{$[A; B]$ indicates vertical concatenation of the matrices A and B.}:
\begin{align}
    \bm w^{(m:l)} &= \begin{bmatrix} \mathrm{PadCol}(\bm w^{(m:l-1)}, d_l); \bm w^{(l)} \end{bmatrix},
\end{align}
where $\mathrm{PadCol}(\cdot, c)$ zero-pads the input matrix to the right by $c$ columns. Then, our final objective from \Cref{eq:relaxed2} with cross-layer dependencies is given as follows:
\begin{align} \label{eq:relaxed3}
\begin{split}
    \mathcal{L}_{train}(\bm w^{(l:L)},\bm s^{(m:l)}, \bm V^{(m:l)})
    =~&\mathcal{L}_{task}\left(f\left(\mathcal{X}; \{\mathrm{sgn}(\bm{w}^{(l)}), \bm{w}^{(l+1:L)}\}\right), \mathcal{Y}\right) \\
    &+ \lambda\left\Vert
        \diag(\bm s^{(m:l)}) {\bm V^{(m:l)}}^\top \left(\bm w^{(m:l)} - \mathrm{sgn}(\bm{w}^{(m:l)})\right)
    \right\Vert_F^2 \\
    &+ \gamma \left\Vert \mathrm{sgn}(\bm{w}^{(m:l)}) \right\Vert_1 + Reg(\bm s^{(m:l)}, \bm V^{(m:l)}).
\end{split}
\end{align}
Given the backbone architecture with $L$ layers, we minimize $\mathcal L_{train}(\bm w^{(l)}, \bm s^{(l)}, \bm V^{(l)})$ with respect to $\bm w^{(l)}, \bm s^{(l)}$, and $\bm V^{(l)}$ to find the desired binarized weights $\bm w_q^{*(l)}$ for layer $l$ while keeping the other layers frozen. Next, we finetune the following layers using the task loss function a few epochs before performing QAT on following layers, as illustrated in \Cref{fig:ours}. This sequential quantization proceeds from the bottom layer to the top and the obtained binarized weights are frozen during the training.

\input{materials/algorithm}

\subsection{Cost-efficient BiTAT via Aggregated Weight Correlation using Reduction Matrix}\label{sec:aggregation}
We derived a QAT formulation which focues on the cross-layer weight dependency by learning block-wise weight correlation matrices.
Yet, as the number of inputs to higher layers is often large, the model constructs higher-dimensional $\bm V^{(l)}$ on upper blocks, which is costly. In order to reduce the training memory footprint as well as the computational complexity, we aggregate the input dimensions into several small groups based on functional similarity using $k$-means clustering. 

First, we take feature vectors, the outputs of the $l$-th layer $\bm{o}^{(l)}_1, \dots, \bm{o}^{(l)}_N \in \mathbb{R}^{d_{l}}$ for each output dimension, to obtain $d_{l}$ points $\bm p_1, \bm p_2, \dots, \bm p_{d_{l}} \in \mathbb{R}^N$, then aim to cluster the points to $k$ groups using $k$-means clustering, each containing $N / k$ points. Let $g_i \in \{1, 2, \dots, k\}$ indicate the group index of $\bm p_i$, for $i = 1, \dots, d_{l}$. We construct the reduction matrix $\bm P \in \mathbb{R}^{k \times d_{l}}$, where $P_{ij}= \frac{1}{N/k}$ if $g_{j} = i$, and $0$ otherwise. Each group corresponds to a single row of the reduced $\widehat{\bm{V}}^{(l+1)} \in\mathbb{R}^{k\times k}$ instead of the original dimension $d_{l}\times d_{l}$. In practice, this \textbf{significantly reduces the memory consumption} of the $\bm V$ (down to \textbf{0.07\%}). 
Now, we replace ${\bm s}$ and ${\bm V}^\top$ in \Cref{eq:relaxed3} to $\widehat{\bm{s}}$ and $\widehat{\bm{V}}^\top \bm P$, respectively, initializing $\widehat{\bm s}$ and $\widehat{\bm V}$ with the grouped input covariance $\frac{1}{N}\sum_{i=1}^N (\bm P\bm  o^{(l)}_i)  (\bm P \bm o^{(l)}_i)^\top$. 
We describe the full training process of our proposed method in \Cref{alg:initial}. The total number of training epochs taken in training is $O(LN_{ep})$, where $L$ is the number of layers, and $N_{ep}$ is the number of epochs for the quantizing step for each layer.

%% file: materials/figures/main_figure.tex
\begin{figure*}
    \small
    \centering
    \begin{tabular}{c}
    \includegraphics[width=1.0\linewidth]{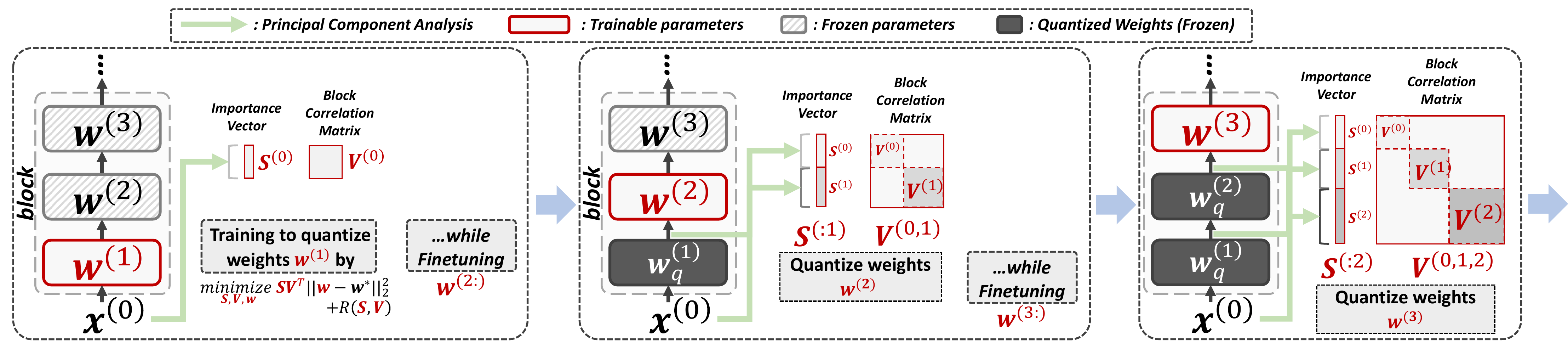}
    \end{tabular}
    \vspace{-0.15in}
    \caption{\textbf{Quantization-aware Training with BiTAT}: We perform a sequential training process: quantization training of a layer - rapid finetuning for upper layers. At each layerwise quantization, we also train the importance vector and orthonormal correlation matrix, which are initialized by PCA components of the current and lower layer inputs in the target block, and guide the quantization to consider the importance of disentangled weights.}
    \label{fig:ours}
\end{figure*}

%% file: materials/figures/4_block.tex
\begin{wrapfigure}{r}{0.25\textwidth}
    \centering
    \small
    \vspace{-.3in}
    \begin{tabular}{c}
    \hspace{-0.15in}
    \includegraphics[height=0.6\linewidth]{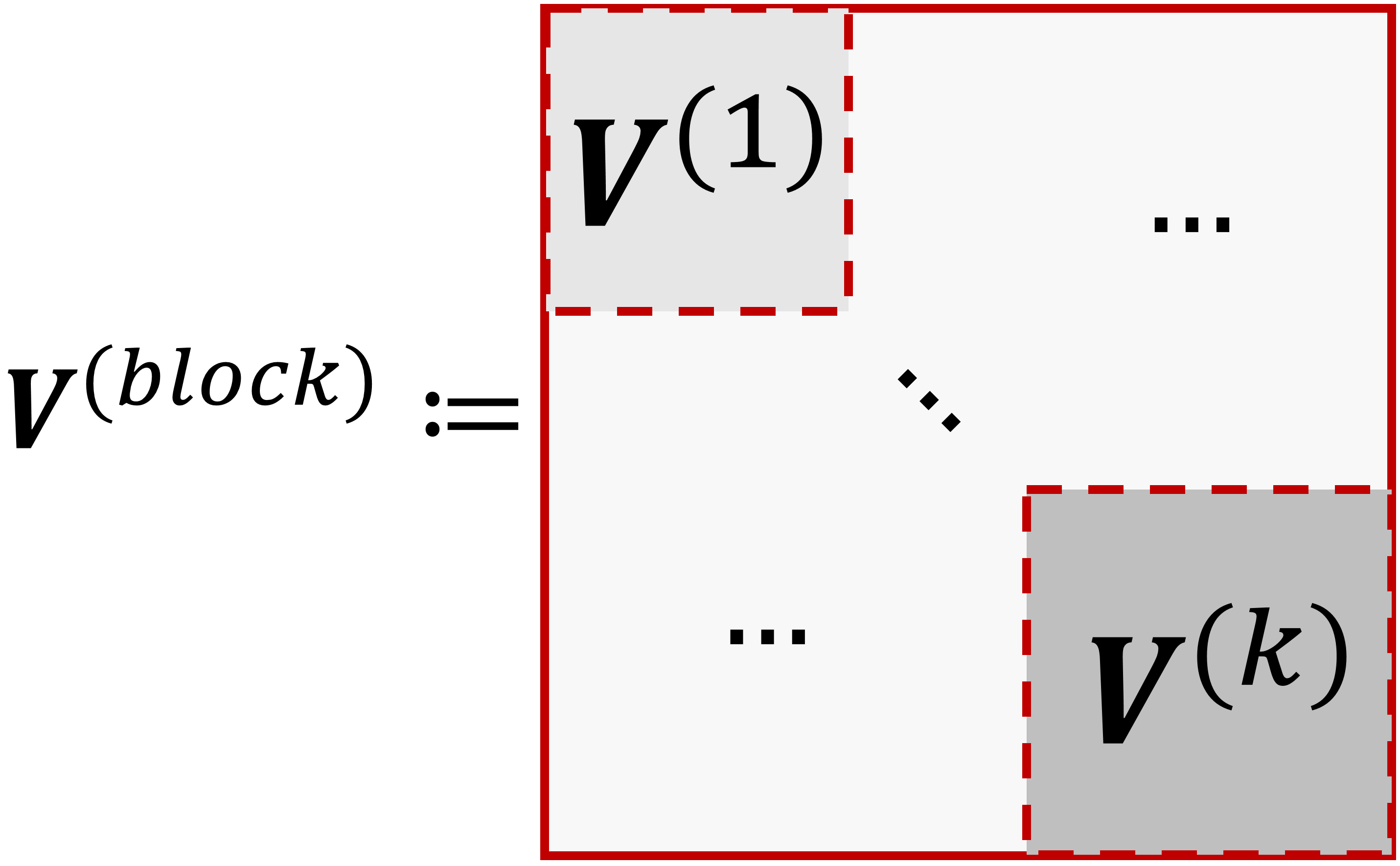}
    \end{tabular}
    \vspace{-0.15in}
    \caption{Initialization of the block correlation matrix.}
    \vspace{-0.25in}
\end{wrapfigure}

%% file: materials/algorithm.tex
\begin{algorithm}[t!]
\small
   \caption{Neural Network Binarization with Task-dependent Aggregated Transformation}
   \label{alg:initial}
\begin{algorithmic}[1]
\STATE {\bfseries Input:} Pre-trained weights $\bm w^{(1)}, \dots, \bm w^{(L)}$ for $L$ layers, task loss function $\mathcal L$, Maximum size of input-dimension group $k$, quantization epochs per layer $N_{ep}$.
\STATE {\bfseries Output:} Quantized weights $\bm w^{*(1)}, \dots, \bm w^{*(L)}$.
\STATE $\mathcal{B}_1, \dots, \mathcal{B}_n$ \textleftarrow Divide the neural network into $n$ blocks
\FOR{{\bfseries each} block $\mathcal B$}
    \STATE $\bm s = [], \bm V = []$
    \FOR{{\bfseries each} layer $l$ in $\mathcal B$}
        \STATE $\bm o^{(l-1)} \leftarrow$ inputs for layer $l$
        \STATE $\bm P \leftarrow$ \textbf{if} $d_{l-1} > k$ \textbf{then} \textsc{K-Means}($\bm X^{(l)}$, $k$) \textbf{else} $\bm{I}_{d_{l-1}}$ \COMMENT{Grouping permutation matrix}
        \STATE $\bm U \diag(\bm \lambda) \bm U^\top = \mathtt{PCA}(\bm P \bm o^{(l-1)})$ \COMMENT{Initialization values \\ for the expanded part}
        \STATE $\bm s \leftarrow [\bm s; \bm \lambda^{\frac 12}]$, $\bm{V} \leftarrow \begin{bmatrix} \bm{V} & \boldsymbol 0 \\ \boldsymbol 0 & \bm{U} \end{bmatrix}$
        \COMMENT{expand $s$ and $V$}
        \STATE $\bm w^{(l:L)}, \bm s, \bm V \leftarrow \argmin_{\bm w, \bm s, \bm V} \mathcal{L}_{train}(\bm w^{(l:L)}, \bm s, \bm V)$ \COMMENT{Iterate for $N_{ep}$ epochs}
        \STATE $\bm w_q^{(l)} \leftarrow \mathrm{sgn}(\bm w^{(l)})$
    \ENDFOR
\ENDFOR
\end{algorithmic}
\end{algorithm}

%% file: sections/5_experiments.tex
\section{Experiments}
\label{subsec:exp_setup}
We validate a new quantization-aware training method, BiTAT, over multiple benchmark datasets; CIFAR-10, CIFAR-100~\cite{krizhevsky2009learning}, and ILSVRC2012 ImageNet~\cite{deng2009imagenet} datasets. We use MobileNet V1~\citep{howard2017mobilenets} backbone network, which is a compact neural architecture designed for mobile devices. We follow overall experimental setups from prior works~\cite{yamamoto_learnable_2021, liu_reactnet_2020}.

\paragraph{Baselines and training details.} While our method aims to solve the QAT problem, we extensively compare our \emph{BiTAT} against various methods; Post-training Quantization (PTQ) method: BRECQ~\cite{li_brecq_2021}, and Quantization-aware Training (QAT) methods: DBQ~\cite{dbq}, EBConv~\cite{bulat_high-capacity_2021}, Bi-Real Net~\cite{liu_bi-real_2018}, Real-to-Bin~\cite{realtobin}, LCQ~\cite{yamamoto_learnable_2021}, MeliusNet~\cite{bethge_meliusnet_2020}, ReActNet~\cite{liu_reactnet_2020}. Note that DBQ, LCQ, and MeliusNet, which keep some crucial layers, such as 1$\times$1 downsampling layers, in full-precision, leading to inefficiency at evaluation time. Due to the page limit, we provide the details on baselines, and the training and inference phase during QAT including hyperparameter setups in the \highlight{Supplementary file}. We also discuss about the limitations and societal impacts of our work in the \highlight{Supplementary file}.

\input{materials/tables/5_main_table}

\subsection{Quantitative Analysis}
\label{subsec:quantitative}
We compare our BiTAT against various PTQ and QAT-based methods in \Cref{tab:main_table} on multiple datasets. 
BRECQ introduces an adaptive PTQ method by focusing on the weight dependency via hessian matrix computations, resulting in significant performance deterioration and excessive training time. DBQ and LCQ suggest QAT methods, but the degree of bitwidth compression for the weights and activations is limited to 2- to 8-bits, which is insufficient to meet our interest in achieving neural network binarization with 1-bits weights and activations. 
MeliusNet only suffers a small accuracy drop, but it has a high OP count. DBQ and LCQ restrict the bit-width compression to be higher at 4 bits so that they cannot enjoy the XNOR-Bitcount optimization for speedup. 
Although Bi-Real Net, Real-to-Bin, and EBConv successfully achieve neural network binarization, over-parameterized ResNet is adopted as backbone networks, resulting in higher OP count. Moreover, except EBConv, these works still suffer from a significant accuracy drop.
ReActNet binarizes all of the weights and activations (except the first and last layer) in compact network architectures while preventing model convergence failure. 
Nevertheless, the method still suffers from considerable performance degeneration of the binarized model.
On the other hand, our BiTAT prevents information loss during quantized training up to 1-bits, showing a superior performance than ReActNet, 0.37 \% $\uparrow$ for ImageNet, 0.53\% $\uparrow$ for CIFAR-10, and 2.31\% $\uparrow$ for CIFAR-100. Note that BiTAT further achieves on par performance of the MobileNet backbone for CIFAR-100. The results support our claim on layer-wise quantization from the bottom layer to the top, reflecting the disentangled weight importance and correlation with the quantized weights at earlier layers.

\paragraph{Ablation study} We conduct ablation studies to analyze the effect of salient components in our proposed method in \Cref{fig:ablation}~\highlight{Left}. 
BiTAT based on layer-wise sequential quantization without weight transformation already surpasses the performance of ReActNet, demonstrating that layer-wise progressive QAT through an implicit reflection of adjusted importance plays a critical role in preserving the pre-trained models during quantization. 
We adopt intra-layer weight transformation using the input-dependent orthonormal matrix, but no significant benefits are observed. Thus, we expect that only disentangling intra-layer weight dependency is insufficient to fully reflect the adjusted importance of each weight due to a binarization of earlier weights/activations. This is evident that BiTAT considering both intra-layer and cross-layer weight dependencies achieves improved performance than the case with only intra-layer dependency. Yet, this requires considerable additional\input{materials/figures/5_matrices}training time to compute with a chunk-wise transformation matrix. In the end, BiTAT with aggregated transformations, which is our full method, outperforms our defective variants in both terms of model performance and training time by drastically removing redundant correlation through reduction matrices. We note that using $k$-means clustering for aggregated correlation is also essential, as another variant, BiTAT with filter-wise transformations, which filter-wisely aggregates the weights instead, results in deteriorated performance.

\input{materials/figures/5_training}
\subsection{Qualitative Analysis}

\paragraph{Visualization of Reduction Matrix} We visualize the weight grouping for BiTAT in \Cref{fig:ablation}~\highlight{Right} to analyze the effect of the reduction matrix, which groups the weight dependencies in each layer based on the similarity between the input dimensions. Each 3$\times$3 square represents a convolutional filter, and each unique color in weight elements represents which group each weight is assigned to, determined by the $k$-means algorithm, as described in \Cref{sec:aggregation}. We observe that weight elements in the same filter do not share their dependencies; rather, on average, they often belong to four-five different weight groups. Opposite to these observations, BRECQ regards the weights in each filter as the same group for computing the dependencies in different layers, which is problematic since weight elements in the same filter can behave differently from each other. 

\paragraph{Visualization of Cross-layer Weight Dependency} In \Cref{fig:matrices}, we visualize learned transformation matrices $\bm V$ (\textit{top row}), which shows that many weight elements at each layer are also dependent on other layer weights as highlighted in darker colors, verifying our initial claim. Further, we provide visualizations for their multiplications with corresponding importance vectors $\diag(\bm s) \bm V^\top$ (\textit{bottom row}). Here, the row of $\bm V^\top$ is sorted by the relative importance in increasing order at each layer. We observe that important weights in a layer affect other layers, demonstrating that cross-layer weight dependency impacts the model performance during quantized training.


%% file: materials/tables/5_main_table.tex
\begin{table*}[t]
\small
\centering
\caption{\small {\bf Performance comparison of BiTAT with baselines.} We report the averaged test accuracy across three independent runs. The best results are highlighted in bold, and results of cost-expensive models ($10^8\uparrow$ ImgNet FLOPs) are de-emphasized in gray. We refer to several results reported from their own papers, denoted as $^\dagger$.\label{tab:main_table}}
\vspace{-0.05in}
\resizebox{\textwidth}{!}{
\begin{tabular}{lccccccc}
\toprule
\textsc{Methods} &
\textsc{Architecture} &
\textsc{\thead{Bitwidth \\ Weight / Activ.}} &
\textsc{\thead{ImgNet\\FLOPs ($\times10^7$)}}&
\textsc{\thead{ImgNet \\ acc (\%)}}&
\textsc{\thead{Cifar-10 \\ acc (\%)}}&
\textsc{\thead{Cifar-100 \\ acc (\%)}}\\
\midrule
\multirow{2}{*}{Full-precision}
& ResNet-18 & 32 / 32 & 
200.0 &
\txtgray{69.8}  &  \txtgray{93.02} & \txtgray{75.61}\\
& MobileNet V2 & 32 / 32 & 
31.40 &
\txtgray{71.9}  & \txtgray{94.43} & \txtgray{68.08}\\
\hdashline
BRECQ~\cite{li_brecq_2021}
& MobileNet V2 & \toxic{4 / 4} & 
~~3.31 &
66.57$^\dagger$ & -  & -\\

DBQ~\cite{dbq}  
& MobileNet V2 & \toxic{4 / 8} & 
~~3.60 &
70.54$^\dagger$  & 93.77 & 73.20\\

\multirow{2}{*}{LCQ~\cite{yamamoto_learnable_2021} }
& ResNet-18 & \toxic{2 / 2} &
15.00 &
\txtgray{68.9}$^\dagger$  & -  & -\\ 
& MobileNet V2 & \toxic{4 / 4} &
~~3.31 &
70.8$^\dagger$  & -  & -\\ 
\hdashline
MeliusNet59~\cite{bethge_meliusnet_2020}
& N/A & \textbf{1 / 1} &
24.50 &
\txtgray{70.7}$^\dagger$  & -  & -\\

Bi-Real Net~\cite{liu_bi-real_2018}
& ResNet-18 & \textbf{1 / 1} & 15.00 & \txtgray{56.4}$^\dagger$  & - & - \\ 

Real-to-Bin~\cite{realtobin}
& ResNet-18 & \textbf{1 / 1} & 15.00 & 65.4$^\dagger$  & - & \txtgray{76.2}$^\dagger$ \\  

EBConv~\cite{bulat_high-capacity_2021}  
& ResNet-18 & \textbf{1 / 1} & 11.00 & \txtgray{71.2}$^\dagger$  & - & \txtgray{76.5}$^\dagger$ \\

ReActNet-C~\cite{liu_reactnet_2020}
& MobileNet V1  & \textbf{1 / 1} &
{14.00} &
\txtgray{71.4}$^\dagger$  &  \txtgray{90.77} & -\\
ReActNet-A~\cite{liu_reactnet_2020}
& MobileNet V1  & \textbf{1 / 1} &
~~\textbf{1.20} &
68.26  &  89.73 & 65.51\\   

\midrule
\bf BiTAT (Ours) & MobileNet V1  & \textbf{1 / 1} &
~~\textbf{1.20} &
\textbf{68.51} &  \textbf{90.21} & \bf 68.36  \\
\bottomrule
\end{tabular}
}
\end{table*}

%% file: materials/figures/5_matrices.tex
\begin{wrapfigure}{o}{0.46\textwidth}
    \vspace{-0.1in}
    \centering
    \includegraphics[width=\linewidth]{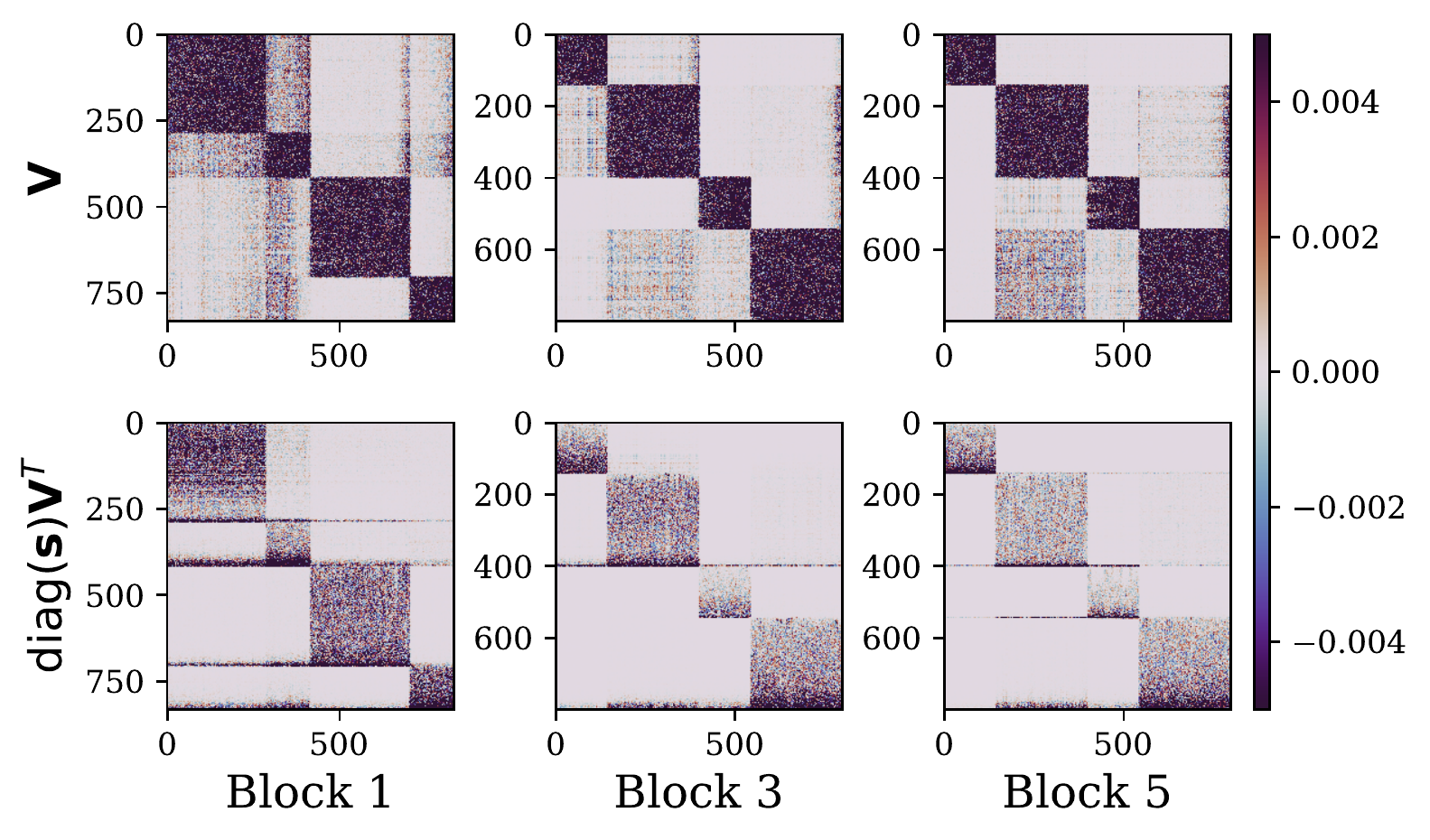}
    \vspace{-0.25in}
    \caption{\small Visualization of the learned $\bm V$ matrix and the $\diag(\bm s)\bm V^\top$ of three blocks of the network, with the CIFAR-100 dataset. Notice the off-diagonal parts which represent cross-layer dependencies. \label{fig:matrices}}
    \vspace{-0.3in}
\end{wrapfigure}

%% file: materials/figures/5_training.tex
\begin{figure*}[t]
\begin{minipage}{0.55\linewidth}
\centering
\resizebox{\textwidth}{!}{
\setlength\tabcolsep{2pt}
\centering
\centering\begin{tabular}{ccccc}
\toprule
\textsc{Method} &\textsc{\thead{Intra-layer\\Transform}} &\textsc{\thead{Cross-layer\\Transform}} &
\thead{Accuracy \\(\%)} &
\thead{Train Time\\(hours)}\\
\midrule
\textsc{ReActNet~\cite{liu_reactnet_2020}} & N/A & N/A & {65.51~$\pm$~0.74} & 10.75\\ 
\midrule
\multirow{4}{*}{\thead{\textsc{BiTAT}\\(Ours)}} 
& $\times$ & $\times$ & {68.17~$\pm$~0.07} &  3.49\\
& $\checkmark$ & $\times$ & {67.82~$\pm$~0.22}  & 3.66\\
& $\checkmark$ & $\checkmark$ & {68.21~$\pm$~0.24} & 8.50\\
&\multicolumn{2}{c}{ \textit{{w/ Filter-wise Transform}}} & {67.86~$\pm$~0.11} & {3.01}\\
&\multicolumn{2}{c}{\cellcolor{gg} \textit{\textbf{w/ Aggregated Transform}}} &\cellcolor{gg} \textbf{68.36~$\pm$~0.45} & \cellcolor{gg} \textbf{3.11}\\
\bottomrule
\end{tabular}}
\end{minipage}
\hspace{-0.2in}
\begin{minipage}{0.5\linewidth}
\centering
\includegraphics[width=0.8\linewidth]{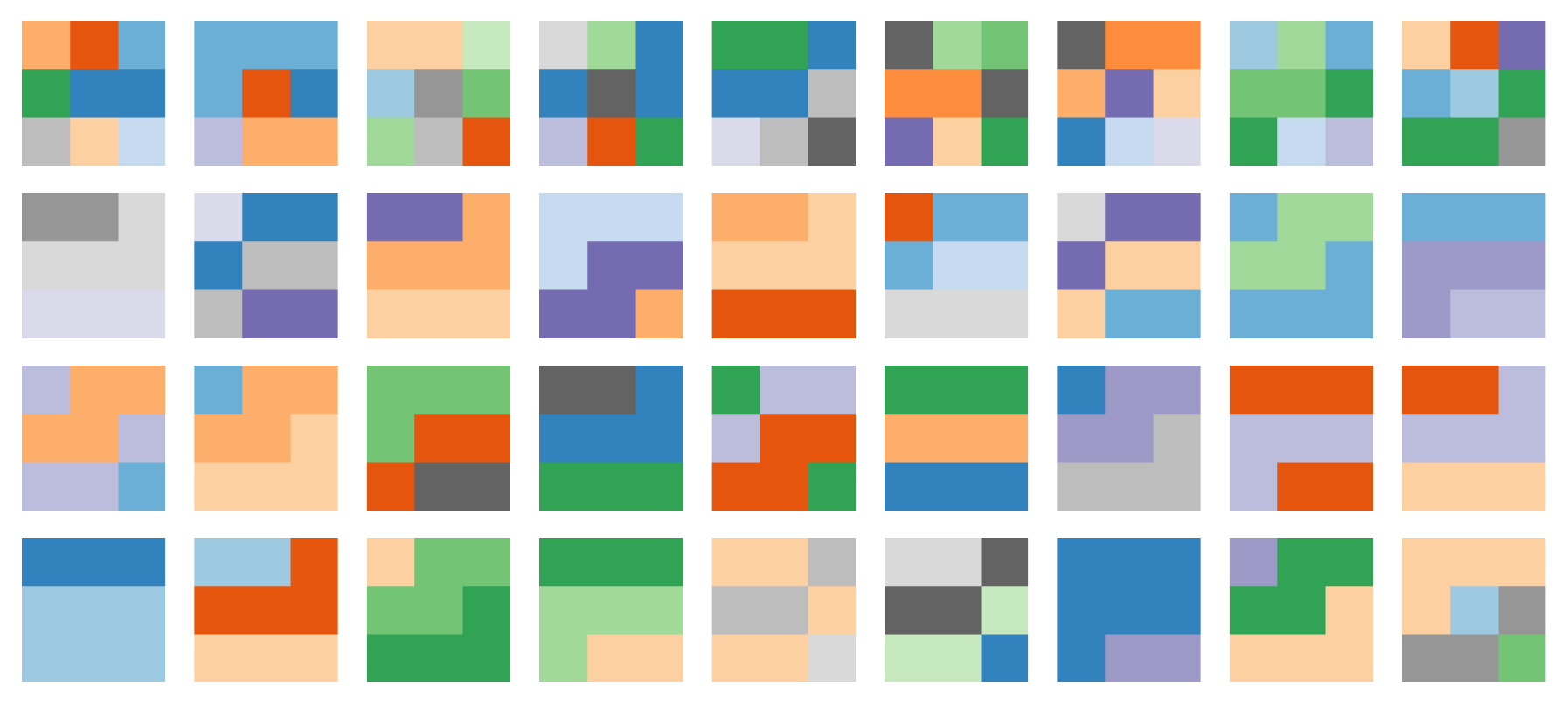}
\end{minipage}
\vspace{-0.05in}
\caption{\small \textbf{Left: Ablation study} for analyzing core components in our method. We report the averaged performance and 95\% confidence interval across 3 independent runs and the complete BiTAT result is highlighted in gray background. \textbf{Right: Visualization of the weight grouping} during sequential quantization of BiTAT. Each 3$\times$3 square represents a convolutional filter of the topmost layer ($26^{th}$, excluding the classifier) of our model, and each unique color represents each group to which weight elements belong.}\label{fig:ablation}
\end{figure*}

\eat{
\begin{figure*}[t]
\hspace{-0.2in}
\begin{minipage}{0.5\linewidth}
\centering
\includegraphics[width=\linewidth]{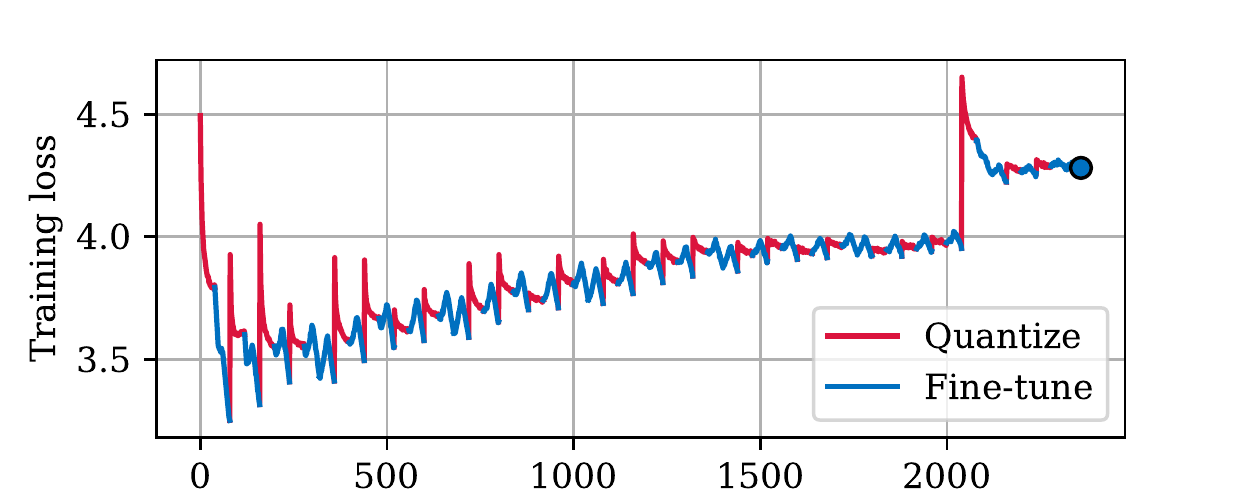}
\end{minipage}
\hspace{-0.25in}
\begin{minipage}{0.57\linewidth}
\centering
\resizebox{\textwidth}{!}{
\setlength\tabcolsep{2pt}
\centering
\centering\begin{tabular}{ccccc}
\toprule
\textsc{Method} &\textsc{\thead{Sequential\\Quantization}} &\textsc{Transform} &
\thead{Accuracy \\(\%)} &
\thead{Train Time\\(hours)}\\
\midrule
\textsc{ReActNet~\cite{liu_reactnet_2020}} & N/A & N/A & {66.05~$\pm$~?.??} & 10.75\\ 
\midrule
\multirow{4}{*}{\thead{\textsc{BiTAT}\\(Ours)}} 
& Layer-wise & $\times$ & {68.17~$\pm$~0.07} &  3.49\\
& Layer-wise & $\checkmark$ & {67.82~$\pm$~0.22}  & 3.66\\
& Block-wise & $\checkmark$ & {68.21~$\pm$~0.24} & 8.50\\
&\cellcolor{gg} Block-wise &\cellcolor{gg} Aggregated &\cellcolor{gg} \textbf{68.36~$\pm$~0.45} & \cellcolor{gg} \textbf{3.11}\\
\bottomrule
\end{tabular}}
\end{minipage}\vspace{-0.1in}
\caption{\small \textbf{Left:} Visualization of the training loss curve during sequential quantization of our BiTAT. \textbf{Right:} Ablation study for analyzing core components in our method. We report the averaged performance and 95\% confidence interval across 3 independent runs and the complete BiTAT result is highlighted in gray background.}\label{fig:ablation}
\vspace{-0.2in}
\end{figure*} 
}

%% file: sections/6_conclusion.tex
\vspace{-.1in}
\section{Conclusion}
In this work, we explored long-overlooked factors that are crucial in preventing the performance degeneration with extreme neural network quantization: the inter-weight dependencies. That is, quantization of a set of weights affect the weights for other neurons within each layer, as well as weights in consecutive layers. Grounded by the empirical analyses of the node interdependency, we propose a Quantization-Aware Training (QAT) method for binarizing the weights and activations of a given neural network with minimal loss of performance. Specifically, we proposed orthonormal transformation of the weights at each layer to disentangle the correlation among the weights to minimize the negative impact of quantization on other weights. Further, we learned scaling term to allow varying degree of quantization error for each weight based on their measured importance, for layer-wise quantization. Then we proposed an iterative algorithm to perform the layerwise quantization in a progressive manner. We demonstrate the effectiveness of our method in neural network binarization on multiple benchmark datasets with compact backbone networks, largely outperforming state-of-the-art baselines.


%% file: sections/7_appendix.tex
\newpage
\section{Appendix}\label{appendix}

\subsection{Details for Problem Setups}
\paragraph{Baselines.} While our method aims to solve the QAT problem, we extensively compare our \emph{BiTAT} against various Post-training Quantization (PTQ)- or QAT methods:
BRECQ~\cite{li_brecq_2021} is a PTQ method that considers weight dependencies using the Hessian of the task loss. DBQ~\cite{dbq} is a QAT method based on continuous relaxation of the quantizer function. EBConv~\cite{bulat_high-capacity_2021} conditionally selects appropriate binarized weights based on the task information. Bi-Real Net~\cite{liu_bi-real_2018} adds residual connections to propagate full-precision values, preventing information loss to activation quantization. Real-to-Bin~\cite{realtobin} constrains a loss term at the end of each convolution to minimize the output discrepancy between the full-precision and the quantized model. LCQ~\cite{yamamoto_learnable_2021} devises a trainable quantization function in order to reduce the quantization error. MeliusNet~\cite{bethge_meliusnet_2020} proposes a new architecture that better propagates full-precision values throughout the network. ReActNet~\cite{liu_reactnet_2020} is the state-of-the-art binary quantization method, which additionally adopts residual connections, and element-wise shift operations before/after the activation and the sign operation. Note that DBQ, LCQ, and MeliusNet keep some crucial layers of MobileNet in full-precision, leading to inefficiency at evaluation time. 

\paragraph{Training.}
Following the setup from ReActNet~\cite{liu_reactnet_2020}, we quantize all layers' weights and activations except the initial and final layers. We use the Adam optimizer~\cite{kingma_adam_2017}. For the ImageNet experiment, a learning rate is $0.002$ and $0.0002$ for quantization training and the fine-tuning, respectively, with linear learning rate decay. We set batch size as $512$ both the quantization phase and the fine-tuning phase is done for $5$ epochs per layer. For the CIFAR-100 experiment, a learning rate is $3\times 10^{-4}$ for quantization training and the fine-tuning with linear learning rate decay. We set batch size as $800$ and both the quantization and fine-tuning are done with $40$ epochs per layer. For all experiments, we set $\lambda = 100$, and $\gamma = 10^{-5}$, which notes that simple choice of the hyperparameters for our regularization terms is sufficient to show impressive performance. The number of input dimension groups is set $k=256$, applying the grouped weight correlation to layers with input dimensions smaller than $k$.

\paragraph{Inference.} In deployment, the highly efficient XNOR-Bitcounting operations can be used for the convolutional layers, also used in existing neural network binarization  works~\cite{courbariaux_binarized_2016, xnornet, liu_reactnet_2020}.

\subsection{Extension to Convolutional Layers}
Let us consider a convolutional layer of size $n_{out}\times n_{in}\times k\times k$, where $n_{in}$ and $n_{out}$ are the number of input and output channels, respectively, and $k$ is the kernel size. We define the set $\mathcal{P}_{\bm{x}}$ as the set of all patches of size $n_{in} \times k \times k$ extracted from the training image $\bm{x}$. This patch-extracting operation is sometimes called \texttt{im2col} or \texttt{F.unfold} in PyTorch.

A convolutional layer applied to $\bm x$ can be thought of as a fully-connected layer individually applied to all patches in $\mathcal{P}_{\bm{x}}$ and then concatenated:
\begin{align}
    \bm w * \bm x = \{\mathrm{Reshape}_{(n_{in}k^2)\times(n_{out})}(\bm w)^\top \mathrm{vec}(\bm p)\}_{\bm p \in \mathcal{P}_{\bm{x}}},
\end{align}
where $*$ denotes the convolution operation, $\mathrm{Reshape}_{shape}(\cdot)$ denotes the reshaping of the tensor into the specified \emph{shape}, and $\mathrm{vec}(\cdot)$ denotes the flattening operation. Each pixel of the output feature map corresponds to a matrix multiplication between a patch and the weight matrix. Therefore, we can analogically apply the same transformation as explained in \highlight{Section 3} to convolutional layers. 

\subsection{Details on Cross-layer Dependency}
In this section, we further explain the detailed experimental setting for \Cref{fig:motiv}. We take the standard MobileNetV2~\cite{sandler_mobilenetv2_2019} model and train it to convergence on the CIFAR-100 dataset with standard SGD with a weight decay. Then, we add noise to the same pretrained model parameters before evaluating the test accuracy based on the following two different ways:
\begin{enumerate}
    \item \textbf{Layer-dependent noise addition.} We first compute the covariance of the input to the first layer and perform PCA using obtained covariance values to compute $\widetilde{\bm{w}}^{(1)}$ in \Cref{eq:tilde}. Now, we add independent gaussian noise with varying scales to the top five rows of $\widetilde{\bm{w}}^{(1)}$. 
    Next, we sequentially repeat the process to the consecutive layers, and after that, we evaluate the model performance, which is shown in solid red lines. The same process is done but with the bottom five rows of each layer instead of the top five, shown in solid blue lines.
    \item \textbf{Layer-independent noise addition.} Before adding noise to model parameters, we compute the covariance of the input values for all layers. Next, we perform PCA and compute $\widetilde{\bm{w}}^{(l)}$ with \Cref{eq:tilde} per layer using these initial covariance values. That is, a layer cannot reflect the weight change through noise addition in others layers, as different from the first approach. Independent gaussian noise with varying scales is added to the top five rows of $\widetilde{\bm{w}}^{(l)}$ for each layer, and then the performance of the model is evaluated, shown in red dashed lines. The same is done with the bottom five rows of $\widetilde{\bm{w}}^{(l)}$, shown in blue dashed lines.
\end{enumerate}

\subsection{Additional Analysis}
\input{materials/figures/additional}
This paper suggests that the proposed quantization loss on disentangled weights is a better indicator for prediction accuracy than the general quantization loss (\Cref{eq:q}), which is evident in multiple validation analyses and the superior model performance in our BiTAT as described in the main text. Here, we provide the quantitative analysis to show that ReActNet~\cite{liu_reactnet_2020} fails to minimize the quantization loss on disentangled weights while our BiTAT successfully does. In \Cref{fig:additional}~\highlight{Left}, we show the $Q_{orig}$ (\Cref{eq:q}) and {$Q_{ours}$} (\Cref{eq:ourq} w/o $\ell_1$ norm) between the initial full-precision weights of the pre-trained model and the obtained binarized weights from ReActNet and BiTAT. $Q_{orig}$ represents the naive MSE between the full-precision weights and the binarized weights. $Q_{ours}$ represents the dependency-weighted MSE between the full-precision weights and the binarized weights.
Note that, in this analysis, we obtain $\bm s$ and $\bm V$ for each layer from the initial pre-trained model by \Cref{eq:initalize} to compute $Q_{ours}$ and neglect the weight dependency across different layers, which is hard to be computed analytically.

We observe that while the value of $Q_{orig}$ is lower in ReActNet than in BiTAT, $Q_{ours}$ is higher in ReActNet than in BiTAT. As shown in \Cref{fig:additional}~\highlight{Right}, the ratio {$r=Q_{ours}/Q_{orig}$} in ReActNet (Red) drastically increases at the beginning stage and is maintained in a high degree until the completion of the quantization-aware training, compared to the $r$ value of the model obtained by BiTAT (Blue dashed). While disregarding the first few epochs of ReActNet training, where the accuracy is very low, ReActNet's $r$ value dominates that of BiTAT. The value slowly decreases as the ReActNet training proceeds, but never reaches the level of BiTAT, demonstrating the inefficiency of the ReActNet training procedure compared to ours.

\subsection{Limitations}
We consider two limitations of our work in this section. First, our BiTAT framework is built based on a sequential quantization strategy, which progressively quantizes the layers from the bottom to the top. Therefore, the training time of our algorithm depends on the number of layers in the backbone network architecture. While we have already validated the cost-efficiency of our proposed method against ReActNet using MobileNet ($26$ stacked layers) in \highlight{Figure 7 Left}, we might spend more training time quantizing all layer weights for the backbones, composed of much more layers like ResNet-$1001$ ($1001$ stacked layers). Next, our method focuses on the cross-layer weight dependency within each neural block, including a few consecutive layers. We thereby avoid the excessive computational cost of obtaining the relationship across all layers in the backbone architecture, yet we consider it a tradeoff between accurate dependency and computation budgets.

%% file: materials/figures/additional.tex
\begin{figure*}
\centering
\begin{minipage}{0.45\linewidth}
\centering
\resizebox{\textwidth}{!}{
\setlength\tabcolsep{2pt}
\centering
\centering\begin{tabular}{cccc}
\toprule
\textsc{~~~~Method~~~~} &
\thead{~~~~Accuracy (\%)~~~~} &
\thead{~~~~$Q_{orig}$~~~~} &
\thead{~~~~$Q_{ours}$~~~~}\\
\midrule
\textsc{ReActNet~\cite{liu_reactnet_2020}} & {65.51~$\pm$~0.74} & \textbf{13.35} & {475.94} \\ 
\thead{\textsc{BiTAT} (Ours)} & \textbf{68.36~$\pm$~0.45} & 39.77 & \textbf{434.92} \\
\bottomrule
\end{tabular}}
\end{minipage}
\hspace{-0.2in}
\begin{minipage}{0.5\linewidth}
\centering
\includegraphics[width=0.8\linewidth]{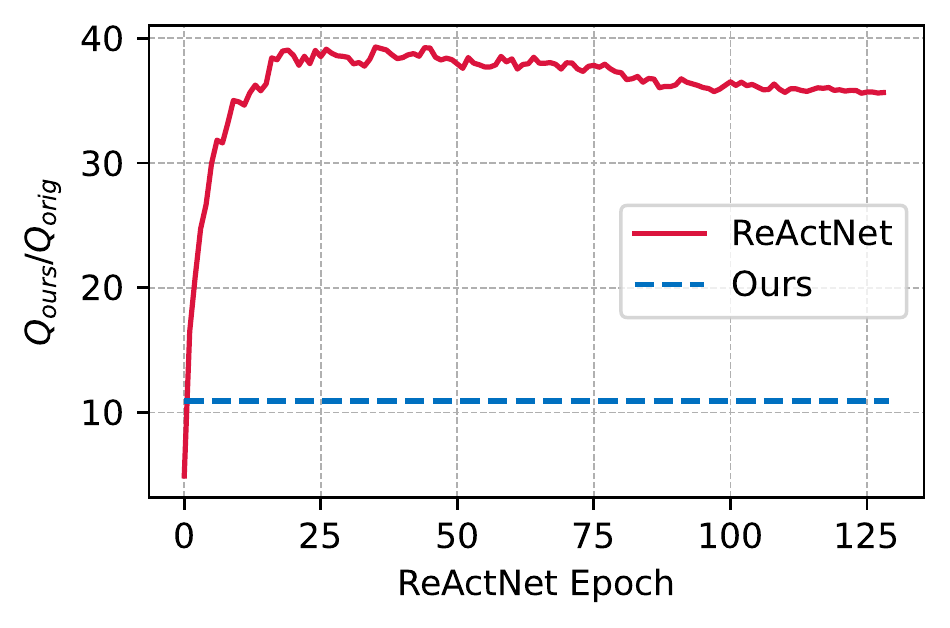}
\end{minipage}
\vspace{-0.1in}
\caption{\small \textbf{Additional ablation studies.} \textbf{Left:} The comparison of the final $Q_{orig}$ (\Cref{eq:q}) and $Q_{ours}$ (\Cref{eq:ourq}) values in ReActNet and BiTAT. \textbf{Right:} The evolution of the ratio of $Q_{ours}$ to $Q_{orig}$ in ReActNet, in comparison to the final ratio in our BiTAT. Cross-layer dependencies not considered in both computations.}\label{fig:additional}
\end{figure*} 